\documentclass{article}
\usepackage{arxiv}
\usepackage[utf8]{inputenc} 
\usepackage[T1]{fontenc}    
\usepackage{hyperref}       
\usepackage{url}            
\usepackage{booktabs}       
\usepackage{amsfonts}       
\usepackage{nicefrac}       
\usepackage{microtype}      
\usepackage{lipsum}
\usepackage{times}
\usepackage{helvet}
\usepackage{courier}  
\usepackage{caption}
\usepackage{soul}
\usepackage{amsmath,bm}
\usepackage{amsfonts}
\usepackage{amssymb}
\usepackage{graphicx}
\usepackage{tabularx}
\usepackage{color}
\usepackage{xcolor}
\usepackage{multicol}

\newcommand{\mbfx}{\mathbf{x}}
\newcommand{\mbfa}{\mathbf{a}}

\newcommand{\mbfy}{\mathbf{y}}
\newcommand{\mbfc}{\mathbf{c}}

\newcommand{\mbfd}{\mathbf{d}}
\newcommand{\mbff}{\mathbf{f}}

\usepackage{letltxmacro}
\LetLtxMacro\oldttfamily\ttfamily
\DeclareRobustCommand{\ttfamily}{\oldttfamily\csname ttsize\endcsname}
\newcommand{\setttsize}[1]{\def\ttsize{#1}}%
\setttsize{\small}

\title{A Knowledge Representation Approach to Automated \\Mathematical Modelling}

\author{
  Bahadorreza Ofoghi \\
  School of Information Technology\\
  Deakin University\\
  Burwood, VIC 3125 Australia \\
  \texttt{b.ofoghi@deakin.edu.au} \\
   \And
  Vicky Mak-Hau \\
  School of Information Technology\\
  Deakin University\\
  Burwood, VIC 3125 Australia \\
  \texttt{vicky.mak@deakin.edu.au} \\
   \And
  John Yearwood \\
  School of Information Technology\\
  Deakin University\\
  Burwood, VIC 3125 Australia \\
  \texttt{john.yearwood@deakin.edu.au} \\
}

\begin{document}
\maketitle
\begin{abstract}
In this paper, we propose a new mixed-integer linear programming (MILP) model ontology and a novel constraint typology of MILP formulations. MILP is a commonly used mathematical programming technique for modelling and solving real-life scheduling, routing, planning, resource allocation, and timetabling optimization problems providing optimized business solutions for industry sectors such as manufacturing, agriculture, defence, healthcare, medicine, energy, finance, and transportation. Despite the numerous real-life Combinatorial Optimization Problems found and solved and millions yet to be discovered and formulated, the number of types of constraints (the building blocks of a MILP) is relatively small. In the search for a suitable machine-readable knowledge representation structure for MILPs, we propose an optimization modelling tree built based upon an MILP model ontology that can be used as a guide for automated systems to elicit an MILP model from end-users on their combinatorial business optimization problems. Our ultimate aim is to develop a machine-readable knowledge representation for MILP that allows us to map an end-user's natural language description of the business optimization problem to an MILP formal specification as a first step towards automated mathematical modelling.

\end{abstract}


\section{Introduction} 
Combinatorial Optimization Problems (COPs) arise in a large number of real-life applications, such as scheduling \cite{Mak2021,Velez2013}, planning \cite{Akartunali2015,Kruijff2018}, resource allocation \cite{Lalbakhsh2018,Mak2017}, routing \cite{Mak2007,Seixas2013}, and time-tabling \cite{Ghoniem2016,Zhou2020}. See, for example,  \cite{chen2010,Gorman2020,Williams2013} for more examples of mathematical programming applications in real-life business COPs where massive cost cuts were achieved. 

There are several commonly-employed solution approaches for COPs. The two main branches are exact methods and heuristic methods. Mathematical Programming is an exact method that can provide proven optimal solutions, and even when it fails to produce an optimal solution within a predetermined time and memory limit, it can still provide a proven optimality gap. Heuristic approaches (such as trial-and-error, simulation, learning, meta-heuristic or custom-made problem-specific heuristic) on the other hand, do not have a solution guarantee. With meta-heuristics or learning methods, with parameters properly tuned, they may be able to provide reasonably good quality solutions within a much shorter time. Exact algorithms will always be preferred in applications where a proven optimal solution matter. For instance, in Kidney Exchange Optimization, an increment of one unit in the objective function means one more kidney transplant can be carried out, and undoubtably will have a significant impact on the health outcome of the patient with a kidney failure. 

When solving a COP, mathematical Programming-based exact algorithms essentially implement an exhaustive tree search with smart pruning strategies. In the case of Integer Programming (IP)-family of methods, the theoretical basis is algebra, whereas in the case of Constraint Programming (CP), the theoretical basis is logical inferences. CP and IP each has their strengths and weaknesses. IP-family of methods include Pure Integer Programming where all decision variables are integers, Binary Integer Programming where all decision variables are binary, and Mixed-integer Linear Programming (MILP) where some decision variables are continuous, and the rest are binary or general integers. MILP can also model some nonlinear terms (e.g., quadratic, bilinear, and piecewise linear terms), and therefore MILP is a very practical technique in modelling and solving real-life COPs. This is evidenced by the fact that in the history of Franz Edelman Awards, 20\% of the finalists applied IP-family of methods \cite{Gorman2020}, and that the top two algorithms in the 2$^{nd}$ Nurse Rostering Competition are MILP-based methods \cite{Ceschia2019}. 

The formal mathematical specification of a MILP problem is given as follows: 
$\{\min, \ \max\}$    
$\{ \mbfc \cdot \mbfx  + \mbfd \cdot \mbfy \ : \ A\mbfx + B\mbfy \leq \mbff, \ \mbfx \in \mathbb{Z}^n_+, \ \mbfy \in \mathbb{Q}^p_+\}$, 
with $\mbfx=(x_1,\ldots,x_n)$ and $\mbfy=(y_1,\ldots,y_p)$ the decision variables; $\mbfc$ and $\mbfd$ the cost coefficients, for $\mbfc$ a $n$-vector of $\mathbb{Q}$, $\mbfd$ a $p$-vector of $\mathbb{Q}$; $A \in \mathbb{Z}_{m \times n}$ and $B \in \mathbb{Q}_{m \times p}$ the constraint coefficient matrices; and $\mbff$ a $m$-vector of $\mathbb{Q}$. For a thorough exposition of MILP, see, for example, \cite{Nemhauser1998,Pochet2006}. 

\subsection{Related work}
Current automation in mathematical modelling is mainly focused around automatic model selection, e.g.,~\cite{Kotthoff2019,NIPS2015_5872} and Learning by Examples (solutions), e.g.,~\cite{dmcp_2016,Pawlak2017}. However, in the context of generalized mathematical modelling, these methods fall short as they are mainly concerned with parameterized model families. 

There has been previous work in the area of constraint or model acquisition for Constraint Programming problems~\cite{dmcp_2016,BESSIERE2017315} and in model acquisition for Integer Linear Programming~\cite{Pawlak2017}. However, the test cases in these studies are of small scales and in some cases, knowledge about the full set of feasible solutions is required in advance. The (automated) task of constraint or model acquisition based on passive or active learning from positive versus negative examples can be achieved by exhaustive testing within a list of candidate constraints or through the application of logic constraint inference from historical data~\cite{Kumar_2019}. Neither of these approaches is suitable for MILP problems. The reason is, in most cases, end-users do not have access to any feasible solutions (positive examples) or infeasible solutions (negative examples)--even determining the existence of positive examples in most real-life CO problems is NP-complete (e.g., constrained routing problems). In the absence of a full set of feasible solutions, invalid constraints are likely to be reached. Even with the full set of feasible solutions, inferring a MILP model implies finding a complete polyhedral description of solution points which results in a computation time that is exponential to the number of variables. 

The task of automatic model selection and machine learning has similarities with constraint acquisition for CP problems considering the fact that given a set of historical data, several models can be tested with reference to a known evaluation schema~\cite{Steinruecken2019}. The works in the domain of automatic model selection have been mainly concerned with parametrized model families. One recent example is Auto-Weka 2.0~\cite{Kotthoff2019} that employs a Bayesian optimization technique to search through the space of Weka's machine learning algorithms and their relevant hyperparameters to select the best-performing model given a specific data set. Auto-sklearn~\cite{NIPS2015_5872} also makes use of Bayesian optimization to implement a similar model search and selection technique. Auto-sklearn, however, employs additional ensemble construction techniques and meta-learning. The latter is concerned with the utilization of meta-knowledge to find an effective mapping between  problem characteristics and algorithm performances~\cite{Brazdil2008}. 

Automatic solution of Mathematical Word Problems (MWPs) is another closely related domain to the task of automated MILP modelling. Proposed methods to solve automatic MWPs include rule-based (symbolic) artificial intelligence techniques, feature engineering and selection using machine learning techniques, and deep learning~\cite{Zhang2019}. These methods, however, approach the solution of MWPs by starting from a pre-existing textual description of the problem, hence word problems. While this assumption works well for applications such as those with educational purposes, in the context of real-world business solution offering the full description of the problems is not a given. Instead, problem elicitation via interaction with end-users will result in the problem description. 

The above challenges and shortcomings in automated mathematical modelling using learning by examples, automatic model selection techniques, and automatic MWP solution necessitates the development of a knowledge-based semantic approach that can well generalize at least to a specific group of mathematical constraints and solutions within the sub-domain of COPs and MILP models. These models are explained in more detail in the next section. 

The OntoMathPro ontology~\cite{Nevzorova2014} is an OWL-based resource that encapsulates a large number of mathematical concepts and the semantic relationships among them. The ontology covers a wide range in the mathematics domain that could benefit novice students to expert mathematicians for information extraction as well as learning and semantic search of formulas. Mocassin Ontology~\cite{SolovyevZ11} is another ontology that has a focus on mathematical information extraction from scholarly articles in the domain, with a focus on the structure of the mathematics-related articles rather than the abstract mathematical concepts and relations.

OMDoc~\cite{Kohlhase:OMDoc} is an extensible XML-based mathematical terminology and language that implements mathematical modelling markups at the three levels of theory, statement, and object. OMDoc makes use of MathML~\cite{Carlisle2009} and OpenMath~\cite{openmath} objects and definitions of symbols to specify mathematical notations. The OMDoc OWL ontology captures semantic relationships such as whole-part, logical dependency, and verbalizing properties, the latter of which are specifically used to reference other OMDoc elements and provide inline definitions of phrase-level constructs. 

Natural Sciences and Technology~\cite{DobrovL06} is an ontology of mathematical concepts and relations developed in the Russian language, especially covering high-school and novice university-level mathematics. This ontology has a specific focus on information retrieval and text analysis in the mathematics domain and implements relations such as is-a and whole-part as well as dependence or associations. 

ScienceWISE~\cite{sciencewise} has put together a series of scientific articles and concepts in several domains, including mathematics, that are related to each other via a set of ontological relations. This ontology was initially formed on the basis of information available from online encyclopedias and other domain-specific resources. The relations that connect ScienceWISE concepts are both generic/taxonomic (such as is-a and part-whole) and domain-specific (e.g., is a model of). Cambridge Mathematical Thesaurus~\cite{ThomasR2004} is another such ontology, which mostly covers mathematical concepts at the undergraduate level with relations such as dependence and associations. With a focus on education, this ontology has entries in several languages.

\subsection{Outline and contributions} 
Our approach to automatic formulation of COPs as MILP models is to mimic the steps of a human modeller. A human modeller does not have to know every single MILP model ever formulated in the MILP history. What they do need, however, is the knowledge of the ``building blocks'', a number of commonly-used requirement (constraint) types, and the knowledge of how to implement these building-block constraints flexibly and intelligently. In this work, we focus on the knowledge representation task for these MILP building-blocks. In particular, our contributions include the following:

\begin{itemize}
    \item The development of the first MILP model ontology that expresses mathematical expert knowledge. 
    \item The development of an optimization modelling tree (OMT) that contains the commonly-used building-block MILP constraint types.
\end{itemize}

The rest of the paper is organized as follows. In the next section, the new MILP model ontology will be introduced, this is then followed by the descriptions of the constraint types in Section 3. Then, in Section 4, we present the OMT together with a few examples. In the last section, we present some summary remarks and future research directions.

\section{The MILP ontology} 
\begin{figure*}[ht]
\begin{center}
\includegraphics{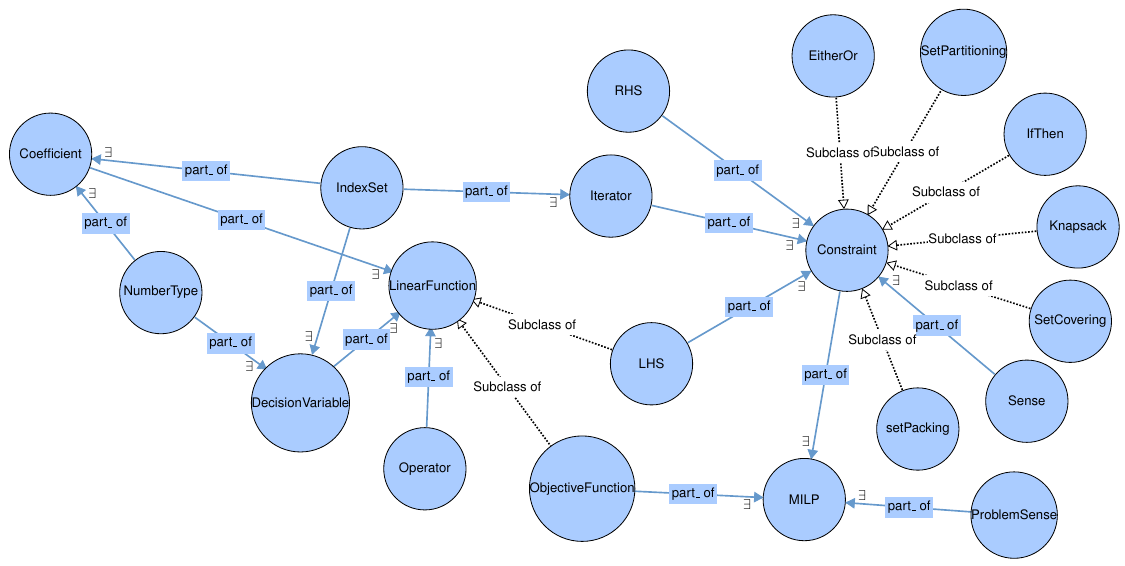}
\caption{The proposed ontology of mixed integer linear programming models for Combinatorial Optimization problems. Note: The visualization is courtesy of WebVOWL~\cite{VOWLpaper}.
}
\label{fig:ontology}
\end{center}
\end{figure*}

We propose the following ontology for MILP models as shown in Figure~\ref{fig:ontology}. The ontology has a top-level class for \texttt{MILP} that has \texttt{Problem Sense}, \texttt{Objective Function}, and \texttt{Constraint} as the main parts. 

An individual instance in the class \texttt{Constraint} has a Left-Hand-Side (LHS) and a Right-Hand-Side (RHS). The LHS is a linear function of decision variables and the RHS is a scalar constant. The \texttt{Linear Function} is a sum of a multiple of coefficients and decision variables. The relevant entities are \texttt{Coefficients}, \texttt{Operators}, and \texttt{Decision Variables}. 
Instances of \texttt{Operator} are $+$ and $\times$. 
Both \texttt{Decision Variable} and \texttt{Coefficient} have \texttt{Number Type} and \texttt{Index Set}. 

A decision variable or a coefficient can have multiple indices and each has a set associated with it. E.g., a set of binary decision variables $x_{i,j,k}$ may have $i = 1,\ldots,10$ that represents the ID numbers of a set of jobs, $j \in J$ for $J$ the set of 12 months in a year, and $k \in K$ the set of staff members. The binary variable may represent whether or not Staff $k$ is to perform Job $i$ in Month $j$. 
Instances of \texttt{Number Type} include $\mathbb{R}_+$, $\mathbb{Z}_+$, $\{0,1\}$. 

The \texttt{Iterator} is similar to the \texttt{Index Set} -- the former enumerates the set of constraints in a MILP and the latter enumerates the set of variables in a MILP. The class \texttt{Problem Sense} has instances $\max$ and $\min$. The class \texttt{Sense} of the constraints has instances $\leq$, $=$, and $\geq$. The \texttt{Objective Function} is also a linear function of decision variables, with ``costs'' of the variables as coefficients. 

The proposed MILP ontology was developed using Protege~\cite{protege} and the Web Ontology Language (OWL). The \texttt{rdfs:subClassOf} relation was used within the OWL representation of the ontology to define the parent-child relations (i.e., is-a) between the different above-mentioned classes, e.g.,

\texttt{<SubClassOf>}\vspace{-5pt}

\texttt{    ~<Class IRI="\#SetCovering"/>}\vspace{-5pt}

\texttt{    ~<Class IRI="\#Constraint"/>}\vspace{-5pt}

\texttt{</SubClassOf>}

\noindent that shows the class \texttt{SetCovering} is a sub class of \texttt{Constraint}.

The part-whole relations (i.e., has-a) are not directly supported within Protege; however, such relations were defined as object properties for the relevant classes;

\texttt{<Declaration>}\vspace{-5pt}

\texttt{    ~<ObjectProperty IRI="\#part\_of"/>}\vspace{-5pt}

\texttt{</Declaration>}
\\

e.g.,\\

\texttt{<SubClassOf>}\vspace{-5pt}

\texttt{    ~<Class IRI="\#Sense"/>}\vspace{-5pt}

\texttt{    ~<ObjectSomeValuesFrom>}\vspace{-5pt}

\texttt{        ~~~<ObjectProperty IRI="\#part\_of"/>}\vspace{-5pt}

\texttt{        ~~~<Class IRI="\#Constraint"/>}\vspace{-5pt}

\texttt{    ~</ObjectSomeValuesFrom>}\vspace{-5pt}

\texttt{</SubClassOf>}

\noindent which expresses the part-whole relation between class \texttt{Sense} and \texttt{Constraint}, the former being a part of the latter.

\section{Constraints} 
\subsection{Constraint type enumeration}
A key question in MILP modelling is whether there is a finite number of MILP constraint types. If so, how many MILP constraint types are there? 
Classic families of COPs such as routing, scheduling, planning have dozens or hundreds of variations from real-life applications. Every COP is different.  Numerous MILPs for real-life COPs found and solved, and many more yet to be discovered and formulated. We are not able to examine every single MILP ever developed in history, however we performed two simple studies for obtaining insights. 1) We examined all constraints used in the production planning problems listed in H. Paul Williams' textbook ``Model building in mathematical programming''. 2) We examined constraints used in a number of publications.

\begin{figure*}[ht]
\begin{center}
\includegraphics[width=16.5cm]{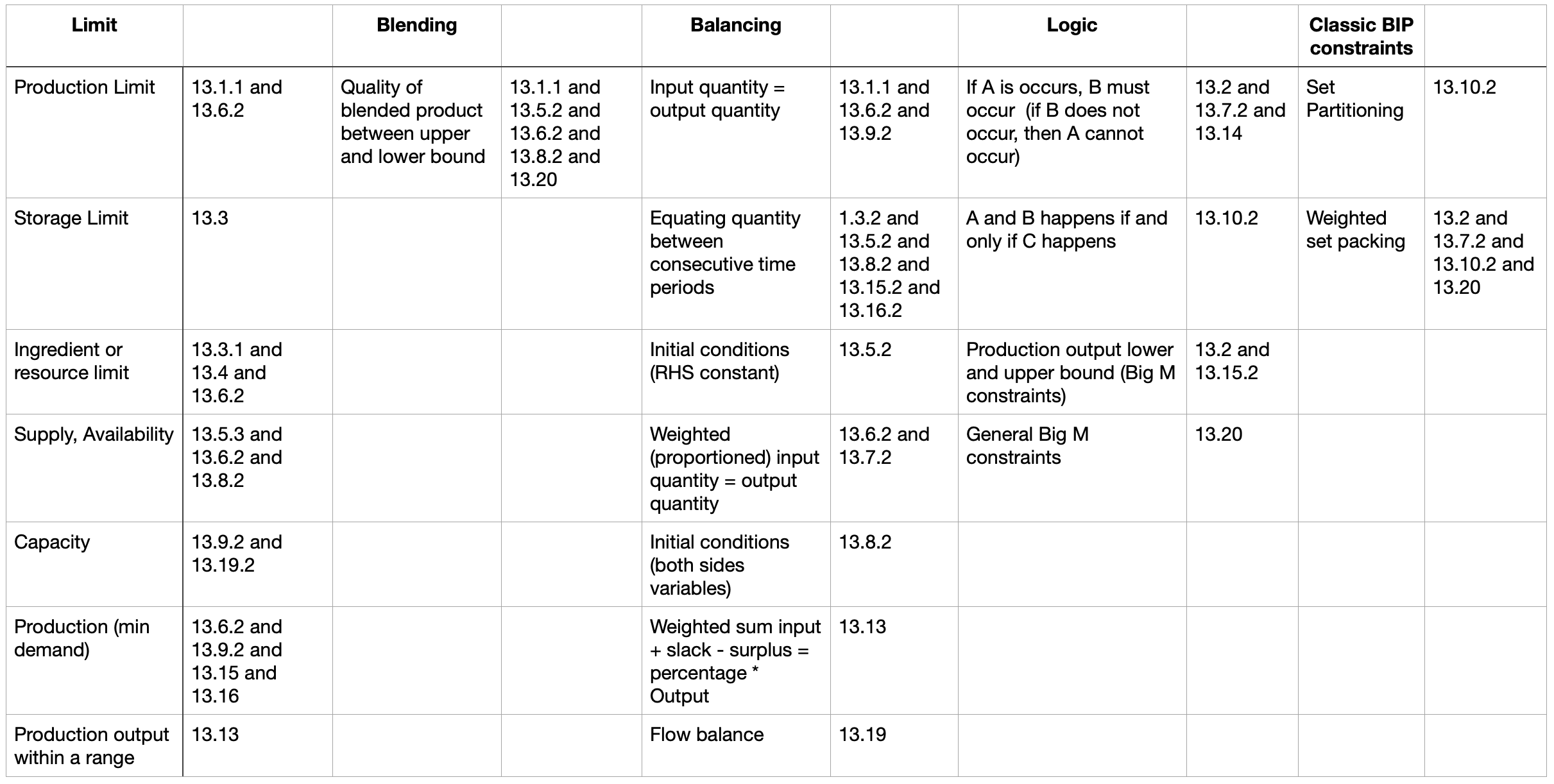}
\caption{A table of constraints and the sections in which they are used in the H Paul Williams book \cite{Williams2013} for production planning problems. 
}
\label{fig:Williams}
\end{center}
\end{figure*}

We considered the production planning examples in H. Paul Williams' book ``Model building in mathematical programming'', and observed that constraints that represent limits (bounds), those that blending from raw materials to products, those that balance two quantities, those that governs logic conditions, and the classic binary integer programming constraints such as set partition, set packing, and set covering and their weighted variations cover all the production planning problems presented therein. In Figure \ref{fig:Williams}, we present a table where we listed the meaning of the constraints used and the section number in the textbook where the examples were discussed. These constraints are reasonably easy to interpret in the sense that the mathematical specification of the constraints is either very close to or can be directly translated from a natural language (NL) description of a n\"{a}ive end-user, (a domain-expert end-user who is not trained to develop MILP models)--we call these explicit constraints.

There are many ways to classify explicit constraints into different types. For instance, if we classify MILP constraint by their mathematical form, they can only be in one of the following forms: $\mbfa \cdot \mbfx \leq b$, $\mbfa \cdot \mbfx = b$, and $\mbfa \cdot \mbfx \geq b$ (here, 
$b\geq 0$). For simplicity, we will write $\mbfa \cdot \mbfx$ as $ax$ for the rest of the paper. 

{\bf Type I: Bound constraints (Demand and Supply)} 
 Resource limit (supply) constraints $ax \leq b$ and demand constraints $ax \geq d$ are very commonly used in MILPs, particularly in production planning-type of problems. For resource limit (supply) constraints, an upper bound on the supply can be fixed (e.g., a Knapsack constraint) or depends on the value of a decision variable. Similar for the lower bound on a  demand that needs to be satisfied. 

{\bf Type II: Balancing constraints} 
Equality constraints $ax = b$ has many variations in its usage: to balance (equate) input and output quantity; to balance the flow or quantities over two consecutive time periods, to set initial conditions, to assign values, and so on. 

{\bf Set packing/partitioning/covering constraints} 
The set packing/partitioning/covering constraints are subtypes of Type I and Type II constraints, typically used for assignment or allocation. They allow us to model the choice of at most/exactly/at least one out of many. The weighted version of the set packing/partitioning/covering constraints allow us to model the choice of $n>1$ out of many. 

{\bf Logic constraints} 
The three main subtypes of {\bf logic constraints} are the {\em Big-M}, {\em If-then}, and {\em Either-or} constraints, each has a number of varieties. (Some of these varieties were discussed in \cite{Ofoghi2020}). 

So, the next question is, what about real-life COPs other than the ones in the \cite{Williams2013} and how do we represent the knowledge in order to enable automatic mapping from business requirements to the mathematical specification (or that of a general purpose modelling language such as OPL or Minizinc)? 

\subsection{Constraint types in MILP formal specifications} 
The mathematics of some of the commonly-used constraint types are presented below. 
In what follows, for simplicity, we use $ax$ to represent the dot product $<\mbfa \cdot \mbfx>$. 
Let $\mbfx \in \{0,1\}^n$ be a set of $n$ binary variables. 
The Set Covering, Set Partitioning, and Set Packing constraints are given by 
\begin{align}
    a  x & \geq   1 \label{SetCoveringConstraint} \\
    a  x & =  1   \label{SetPartitioningConstraint}  \\
    a  x & \leq  1       \label{SetPackingConstraint} 
\end{align}
respectively, for $\mbfa \in  \{0,1\}^n$. ((\ref{SetPartitioningConstraint}) and (\ref{SetPackingConstraint}) are different in modelling though mathematically they are equivalent as one can be transformed to another with the help of additional variables.) 
If the right hand side for (\ref{SetCoveringConstraint}) and (\ref{SetPartitioningConstraint}) are $\mathbb{Z}_+$, then we have the Weighted Set Covering and Weighted Set Partitioning respectively, and if on top of that, $\mbfa \in \{0, 1, -1\}$ then we have the Generalized Set Covering and Generalized Set Partitioning respectively.  
If $\mbfa \in \mathbb{Q}^n_+$, $b \in \mathbb{Z}_+$, and $\mbfx \in \mathbb{Q}^n_+$, then $ax \leq b$ is a Knapsack Constraint. If $\mbfx \in \{0,1\}^n$, we have a 0-1 Knapsack Constraint.  
The following constraints are used for formulating logical relations. 
\begin{itemize}
    \item  {\em Either-or} condition. Suppose we wish to have either $f(\mbfx) \leq 0$ or $g(\mbfx) \leq 0$, we use the constraints $f(\mbfx) \leq Mt$ and $g(\mbfx) \leq M(1-t)$, for $f(\mbfx)$ and $g(\mbfx)$ linear functions of decision variables $\mbfx = (x_1,\ldots,x_n)$, $M$ a sufficiently larger number, and $t$ a binary decision variable. 
    \item {\em If-then} condition. Suppose $g(\mbfx) \leq 0$ will be true if $f(\mbfx) > 0$ is true, we can model such a condition by using  constraints $g(\mbfx) \leq Mt$ and $f(\mbfx) \leq M(1-t)$, again, for $\mbfx = (x_1,\ldots,x_n)$, $M$ a sufficiently large number, and $t$ a binary decision variable. 
\end{itemize}

The {\em if-then} constraints are one of the most commonly used constraint class in modelling logical relations, for example, like the ones below. 
\begin{itemize}
    \item If we produce Product A, then we must also produce at least two of Products B, C, D, or E; 
    \item If a class is not scheduled at a particular time, then no students will be allocated to this time; 
    \item A occurs if and only if all of B, C, D, and E occurs. 
\end{itemize}
A few classic examples are provided below. 

\begin{itemize}
\item 
To model ``if $f(\mbfx)$ occurs ($f(\mbfx) =1$), then $g(\mbfx)$ occurs ($g(\mbfx) =1$)'', for $\mbfx \in \{0,1\}^n$ is a set of binary decision variables, $f(\mbfx)$ and $g(\mbfx)$ linear functions of $\mbfx$ with $f(\mbfx), g(\mbfx) \in \{0,1\}$, we  have that  
\begin{align} 
f(\mbfx) \leq g(\mbfx) \label{If-A-Then-B}
\end{align}
\item 
Let $x_A$ be a binary decision variable with $x_A =1$ if Task $A$ is to be carried out, $x_A =0$ otherwise;  and $x_{B_j}$ for $j = 1,\ldots,n$ be binary variables with $x_{B_j} = 1$ is Task $B_j$ is to be carried out, $x_{B_j} =0$ otherwise. To model $A$ occurs if all of $B_1,\ldots,B_n$ occur, we have that: 
\begin{align} 
x_{B_1} + \cdots+ x_{B_n} \leq n - 1 + x_A \label{IfAthenAllB} 
\end{align}
\item 
For {\em $A$ occurs only if all of $B_1,\ldots,B_n$ occur}, we have that 
\begin{align}
x_A \leq x_{B_j}, \quad \forall j  = 1,\ldots,n \label{AonlyIfAllB} 
 \end{align}
\item 
For {\em $A$ occurs if and only if all of $B_1,\ldots,B_n$ occur}, we use (\ref{IfAthenAllB}) and (\ref{AonlyIfAllB}) together. 
%
%
\item 
{\em If $A$ occurs then $f(\mbfx) = C$, for $C \in \mathbb{Q}$.} We have that: 
\begin{align} 
\label{IfxThenfx=C}
\begin{split}
-M (1-z_A) + f(\mbfx) & \leq  C \\
M(1-z_A) + f(\mbfx) & \geq  C
\end{split}
\end{align}
\end{itemize}


\section{The optimization modelling tree} 
We designed an Optimization Modelling Tree (OMT) and examined a number of COPs in published journal articles to ascertain whether the OMT is adequate in the sense that by traversing through the tree all elements for the MILPs can be found. The focus of this exercise is to evaluate whether the constraint types and subtypes in the OMT are enough to represent these example COPs. 

%
\begin{figure*}[ht]
\begin{center}
\includegraphics[width=16.5cm]{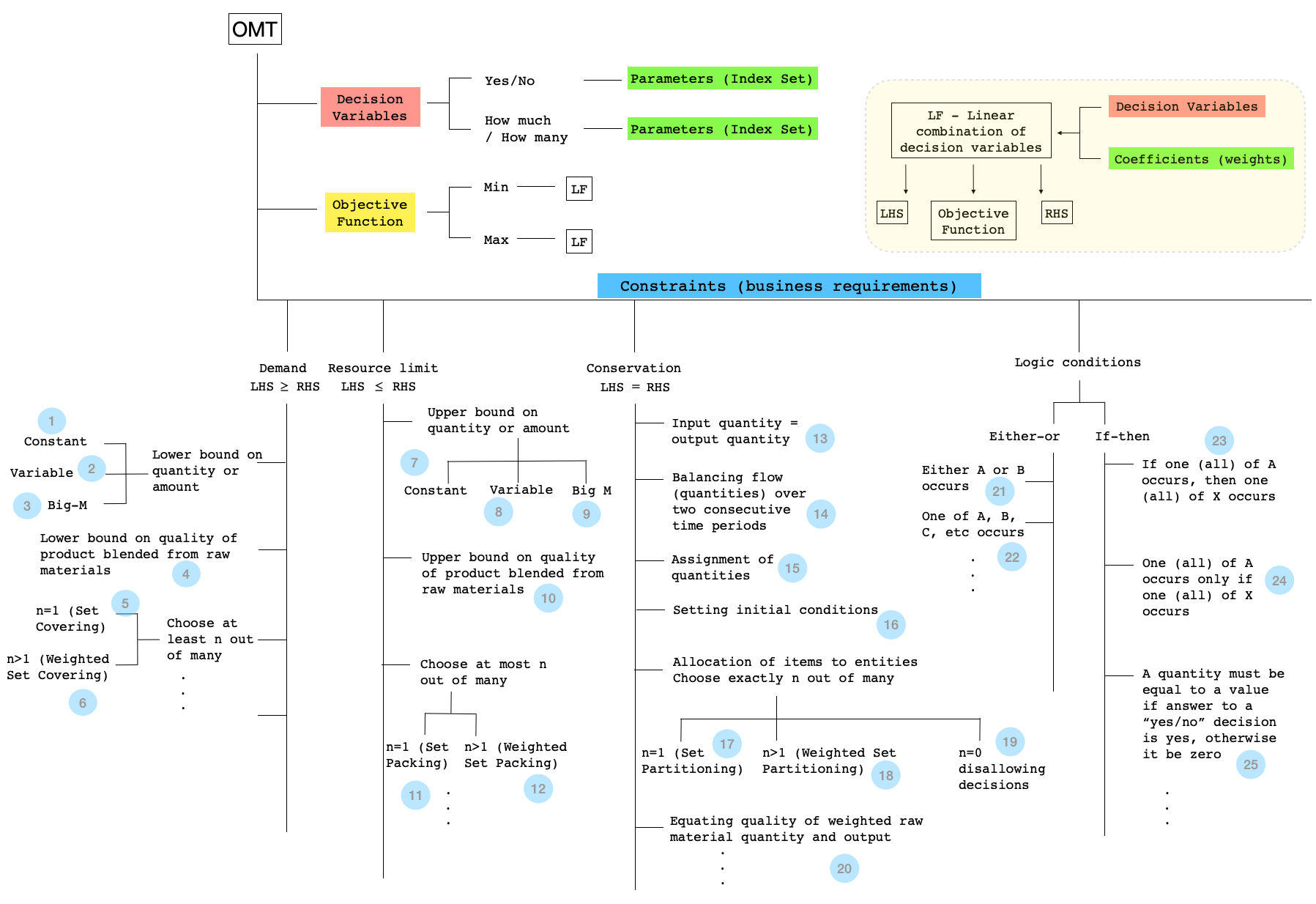}
\caption{The overall structure of the proposed Optimization Modelling Tree (OMT) including the three gerenal classes of decision variables, objective functions, and constraints.
}
\label{fig:OMT}
\end{center}
\end{figure*} 
%

{\bf A chemical production scheduling example} 
A chemical production scheduling MILP model is presented in \cite{Velez2013}. The problem considers a given planning horizon, partitioned into a number of time slots. The decisions to be made are whether a unit (a machine or equipment) should start processing a task at a particular time slot, what the batch size should be, and the inventory level of each material at each time slot. A basic MILP is presented with 4 constraints. Constraint Set (1) is a set packing constraint ensuring that each unit will be starting at most one task at a time (Number 11 on the OMT). Constraint Set (2) is a combined logic (Big-M) and upper/lower bound constraint--if a unit starts processing a task at a given time, then the capacity of the batch size must be observed, otherwise, the task will not be processed on this machine at this time (Numbers 3 and 9 on the OMT). Constraint Set (3) presented in the paper should have been two constraints. One is to equate the inventory (storage) of a material at a time slot to the inventory at the previous time slot plus the new production and minus the  consumption (Number 14 on the OMT). The second part of Constraint Set (3) is an upper bound on the storage limit (Number 7 on the OMT). These constraints are reasonably straight forward to describe by an end-user, and all requirements can be found in the constraint types on the OMT. 

{\bf A supply chains production planning example} 
An MILP model for mid-term production planning for high-tech low-volume supply chains is presented in \cite{Kruijff2018}. The decision variables are mostly general integer variables, and the six constraint sets are as follows. Constraint Set (1) are to balance two quantities, in specific, quantities between two consecutive time slots (Number 12 on the OMT). Constraint Sets (2) and (5) are equality constraints for assigning quantities (Number 13 on the OMT). Constraint Sets (3) and (6) are variable upper bounding and lower bounding (Numbers 2 and 8 on the OMT) whereas Constraint set (4) has the logic condition that the upper and lower bounds on decision variables for quantities apply only when the associated binary decision variables is non-zero (Numbers 3 and 9 on the OMT). 

{\bf A university course timetabling problem example} 
A university course timetabling problem was modelled as an MILP in \cite{Ghoniem2016}. The decision variables are binary. One set of the decision variables represent yes/no answers to whether a particular section of a course should be assigned to a particular professor in a particular time slot. Translating them from NL to formal specifications should be reasonably straight forward.  Constraint sets (2) and (3) are Set Partitioning constraints (for choice of exactly one out of many, Number 17 on the OMT), Constraint Sets (4) to (9) are Set Packing constraints (for choice of at most one out of many, Number 11 on the OMT)). Constraint sets (11)--(13), and (15) are general if-then constraints regulating if $X$ occurs, then both of $Y$ and $Z$ must occur. The constraints are in the form of $2X \leq Y + Z$ (Number 24 on the OMT), although $X \leq Y$ and $X \leq Z$ are better constraints to use.  This brings an important aspect for knowledge engineering MILPs: multiple feasible MILP constraints exist for the same requirement, some are strong for computational use than others. The OMT in its current state has some limitations, as we can see from the next example. 

{\bf A multitrip vehicle routing problem with time windows example}. The COP described in \cite{Seixas2013} is a routing-type problem. An end-user not trained with MILP knowledge does not normally describe that a yes/no decision is associated with each pair of locations (e.g., $i$ and $j$ with a yes answer indicating Location $j$ must be visiting immediate after Location $i$). However, commonly-used MILPs for routing problems typically use a binary variable for each of these decisions. Once the hurdle in decision variable definition is overcome, the rest of the constraints can be found in the constraint types or subtypes described in the OMT. Constraint Sets (1), (3), (4) are all Set Partitioning Constraints, i.e., to choose exactly one out of many (Number 17 on the OMT). Constraint Set (2) is to set to zero variables that represent impossible decisions (Number 19 on the OMT). Constraint Sets (6) to (8) are to regulate the time of arrival of a vehicle route to visit a customer, and the constraints are {\em if-then} subtype be found in the OMT. The last constraint set (10) ensures is a straight forward upper bounding constraint on total time used, and the bound itself is a variables (Number 2 on the OMT). 
Constraint Set (9) is a special type of demand - capacity constraint commonly used in routing-type of problems. The requirement is not trivial to describe in NL by an end-user but the mathematical constraint itself can be found in the OMT (it is in fact a Set Packing Constraint, Number 11 on the OMT). Constraint Set (5) is a flow-balance constraint (which is covered by the OMT), the mathematical meaning is that if a customer is visited, then there must be a customer that was visited before him/her and one after him/her. An end-user would not describe the requirement like this. We call these implicit constraints.

\section{Summary remarks and future research}
In the first attempt to achieve automated mathematical modelling, we have developed an ontology for MILP models. The ontology formulates expert mathematician knowledge in the domain in order to remove (or minimize) the need for the expert in the modelling of MILP-related problems. We have also developed the first Optmization Modelling Tree (OMT) to facilitate automated user requirement elicitation. What the OMT contains is not just the mathematical specification of the MILP constraints but also a configuration of the classes that can assist in the modelling process and be mapped to requirements. Mathematically, $ax \leq b$, $ax = b$, and $ax \geq b$ are enough to cover all MILP constraints. However, the OMT that we designed branches by usage (or the meaning of the constraints in an application). A beginner MILP modeller, for example, can traverse through the tree to elicit business requirements from a non-expert end-user. We have tested some COP instances and found that the constraints in the OMT cover all the ``explicit'' (or straightforward) constraints. Even for constraints in our test cases that are not straightforward, i.e., the ``implicit'' constraints, they too are covered by the OMT, although the mapping mechanism is not represented on the OMT. 

We have the same results for the ACs and the SECs of an ATSP, they can appear in the form of Set Partitioning and Set Covering constraints respectively; however, the mapping is not explicitly represented in the tree. At this stage, we hypothesize that the OMT is scalable; a formal proof is yet to be developed. 


An important further work will focus on the compilation of a list of mappings for commonly-used implicit constraints. For example, the knowledge that ``visiting each city exactly once and return to the home city'' is equivalent to ``each entity has one that precedes it and one that succeeds it'' is needed to be represented on the current OMT and consequently, the Set Partitioning and Set Covering constraints be identified as the right constraints to be used. We are planning to perform a survey of literature for commonly-used implicit constraints and the usage they map to and to represent such a mapping on the OMT in an efficient way.

\bibliographystyle{abbrv}
\bibliography{bibliography}
\end{document}